\documentclass{article}

\PassOptionsToPackage{numbers, compress}{natbib}

\usepackage[preprint]{nips_2018}

\usepackage[utf8]{inputenc} 
\usepackage[T1]{fontenc}    
\usepackage{hyperref}       
\usepackage{url}            
\usepackage{booktabs}       
\usepackage{amsfonts}       
\usepackage{nicefrac}       
\usepackage{microtype}      
\usepackage{graphicx}
\usepackage{subcaption}

\title{Can machine learning identify interesting mathematics? An exploration using empirically observed laws }

\author{
  Chai Wah Wu\\
  IBM Research AI\\
  IBM T. J.~Watson Research Center\\
  P. O. Box 218\\
  Yorktown Heights, NY 10598 \\
  \texttt{cwwu@us.ibm.com} \\
}

\begin{document}
\date{September 9, 2018}

\maketitle

\begin{abstract}
We explore the possibility of using machine learning to identify interesting mathematical structures by using certain quantities that serve as fingerprints. In particular, we extract features from integer sequences using two empirical laws: Benford's law and Taylor's law and experiment with various classifiers to identify whether a sequence is, for example, nice, important, multiplicative, easy to compute or related to primes or palindromes.
\end{abstract}

\section{Introduction}
Machine learning has made significant strides in solving classification problems in several domains that humans excel at, for instance in image processing and speech recognition. There has been some effort to classify scientific knowledge as well \cite{Akritidis2013} by analyzing the text of scientific articles. So far, there has been much less progress in terms of classification using only the mathematical equations and quantities in scientific knowledge. Part of the difficulty is that there is less leeway in the interpretation of mathematics; the same numbers, symbols and equations can have completely different meaning based on the specific way these objects are composed on the page. On the other hand, classical logic-based AI and symbolic computer algebra systems have been more successful in this regard \cite{Kushman2014}. 

The purpose of this paper is to investigate the following perhaps simpler problem: can machine learning identify qualitative attributes of scientific knowledge, i.e. can we tell whether a scientific result is elegant, simple or interesting? We will start our investigation by restricting the domain to mathematical sequences of numbers.

\section{Online Encyclopedia of Integer Sequences}
The object of study in this paper are sequences of integers. One reason for choosing them to study is that many fundamental mathematical ideas are captured in these structures. Another reason is that there exists an extensive database of integer sequences that has been edited and curated for over 50 years: the Online Encyclopedia of Integer Sequences (OEIS) \cite{oeis}. The OEIS was created by Neil Sloane in 1964 and has grown to over 300,000 sequences as of this writing with thousands of volunteers from the OEIS community editing and adding metadata and references to these sequences.
To each sequence are associated keywords assigned by the community members. Some examples of keywords are:`nice', `core', `base', `hard', etc. A complete set of keywords and their definitions can be found at \url{http://oeis.org/wiki/Keywords}.  Many classical sequences are in the database, such as the sequence of primes, the binomial coefficients, the Fibonacci numbers, etc. There is a range of complexity, ranging from sequences that are very easy to compute (such as the sequence of odd numbers \href{https://oeis.org/A005408}{A005408}), hard to compute (such as the number of nonsingular $n\times n$ 0-1 matrices \href{https://oeis.org/A055165}{A055165})  to sequences for which it is not known whether it is finite or not (such as the list of Mersenne primes \href{https://oeis.org/A000668}{A000668}).

\section{Empirical laws}
Since the number of terms of each sequence that are available for analysis can vary, it is desirable to have a fixed number of features that can be computed on sequences of any finite length. An objective of this paper is to study whether empirical laws can serve this purpose. In particular, we look at 2 empirically observed laws that have appeared in the literature. Empirical laws are not mathematical theorems per se, but are empirical observations of relationships that seem to apply to many natural and man-made data sets (e.g. Moore's law in electrical engineering \cite{Moore1965b} or the 80/20 Pareto principle in economics), but why these occur so frequently are typically still not completely understood. As these empirical laws are discovered because many data sets of interest seem to abide by them, they are a good starting point for finding features for classification.

\subsection{Benford's law}
Benford's law \cite{Benford1938} states that in a set of numerical data, terms with a small leading digit tend to occur more frequency. More precisely, in base $b$,
terms with leading digit $d$ occurs with probability equal to $\log_{b}(\frac{d+1}{d})$. In particular we will define the discrete distribution $ \{d_i\}$ where $d_i =  \log_{10}(\frac{i+1}{i})$ and $i=1,\cdots 9$ to be the {\em Benford} distribution $b(i)$.\footnote{This is the base $10$ version of Benford's law which has been verified for many experimental data sets, and it appears to hold in other bases as well.}

This empirical law was first observed by Simon Newcomb \cite{Newcomb1881} who noted that in logarithms tables there were more numbers starting with 1 than with any other digit. This was noted later by Frank Benford who analyzed it for other data sets.
Recently, Benford's law has been shown to apply to several integer sequences \cite{Huerlimann2009}.

\subsection{Taylor's law}
Another empirical law was defined by Lionel Taylor in 1961 \cite{Taylor1961} who noted that in ecology, the mean  $\mu$ and the variance $v$ in species data appear to satisfy a power law:
\begin{equation}
 v = T_a\mu^{T_b}
\label{eqn:taylor}
\end{equation}
where $T_a$ and $T_b$ are positive constants.

Taylor's law has been observed in many naturally observed data sets \cite{Xiao2015,Reuman2017} and in integer sequences such as the list of primes \cite{Cohen2016} and binomial coefficients \cite{Demers2018}.

\section{The data set}
For the data set, we selected 40,000 sequences randomly from OEIS each with at least 990 terms accessible in the database.\footnote{The reason for this odd number (990) is because the terms in the OEIS database are (for the most part) limited to 1000 digits, and some sequences such as ``smallest prime containing at least $n$ consecutive identical digits'' (OEIS sequence \href{https://oeis.org/A034388}{A034388}) will have slightly less than 1000 terms in the database.} On average, approximately 1 in 4 sequences in the OEIS contains over 990 terms in their entry. Although most sequences in OEIS are defined as infinite sequences, it is not always easy to compute many terms. For each sequence, we collect all the terms that are available in OEIS and compute several quantities for each sequence.

\subsection{Checking for Benford's law}
To check for Benford's law, we compute the following features for each sequence:
$b_d(i)$ is the proportion of terms with leading digit $i$ for $i = 1,\cdots 9$.

For each sequence $\{a(n)\}$,\footnote{As mentioned before, since the sequences are generally infinite, we mean here all the terms of the sequence that are available in the OEIS database.} we compared the proportion of terms $b_d(i)$ with the Benford distribution $b(i)$ by using 4 different statistical distances: (1) the Kullback-Leibler (KL) divergence $D_{KL}(b_d||b) = \sum_i b_d(i)\log \frac{b_d(i)}{b(i)}$, (2) the Kolmogorov-Smirnov (KS) statistic $KL(b_d,d)$, (3) the Wasserstein distance (or earth mover's distance) $WD(b_d,b)$ and (4) the total variation distance $TV(b_d,d)$.  

\begin{figure}[htbp]
\begin{subfigure}{0.5\textwidth}
\centerline{\includegraphics[width=2.9in]{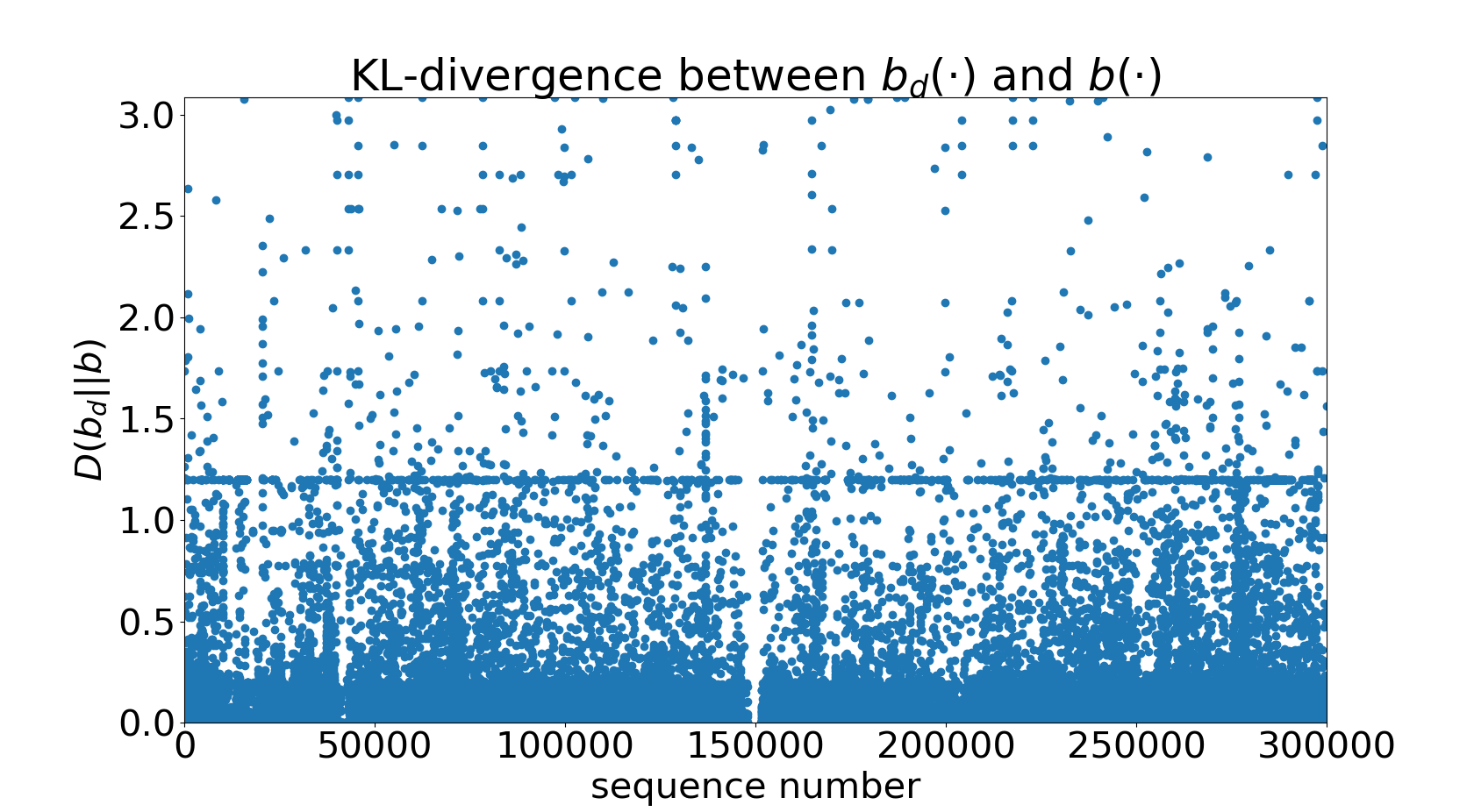}}
\caption{Kullback-Leibler divergence of $b_d$ versus $b$.}
\label{fig:KL}
\end{subfigure}
\begin{subfigure}{0.5\textwidth}
\centerline{\includegraphics[width=2.9in]{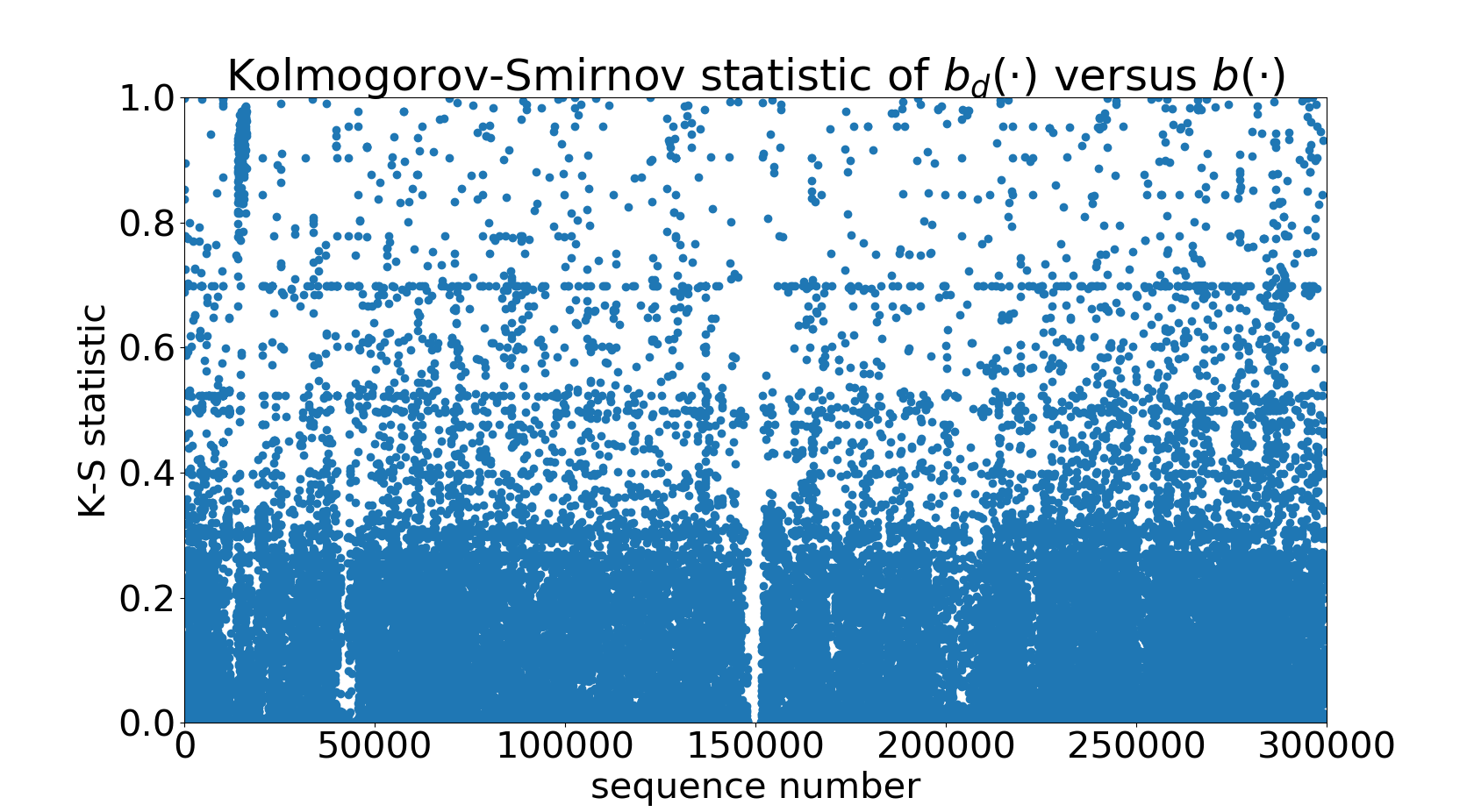}}
\caption{Kolmogorov-Smirnov statistic of $b_d$ versus $b$.}
\label{fig:KS}
\end{subfigure}

\bigskip
\begin{subfigure}{0.5\textwidth}
\centerline{\includegraphics[width=2.9in]{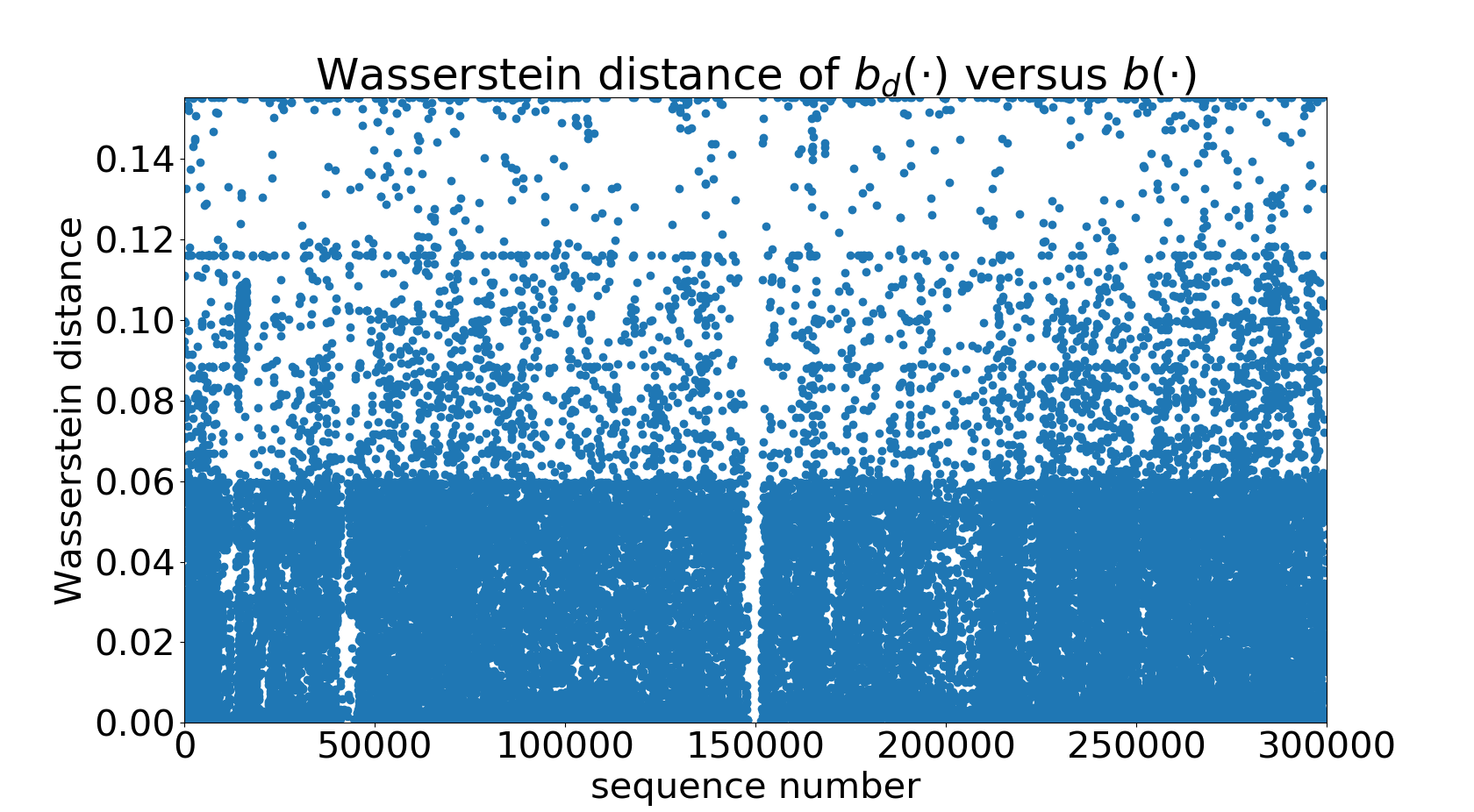}}
\caption{Wasserstein distance of $b_d$ versus $b$.}
\label{fig:WD}
\end{subfigure}
\begin{subfigure}{0.5\textwidth}
\centerline{\includegraphics[width=2.9in]{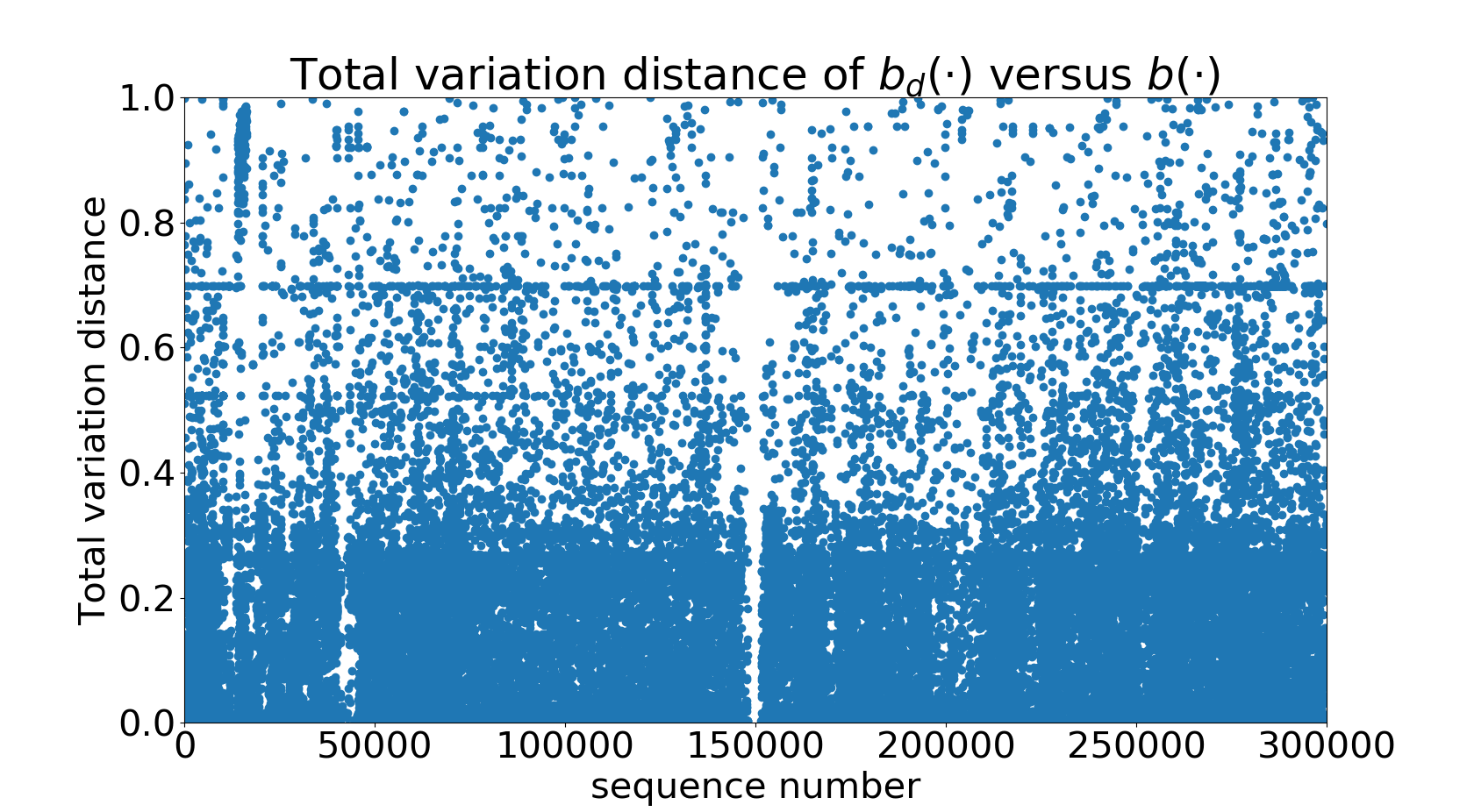}}
\caption{Total variation distance of $b_d$ versus $b$.}
\label{fig:TV}
\end{subfigure}
\caption{$4$ statistical distances between $b_d$ and the benford distribution $b$. }
\label{fig:benford}
\end{figure}

Fig. \ref{fig:KL} shows the KL divergence of the various sequences.
We see that for many of the sequences, the KL-divergence is small and in the range [0,0.2]. The KL-divergence between the uniform distribution and $b(i)$ is 0.191, indicating that for most sequences $b_d(i)$ is a decreasing function. There is a cluster of sequences with KL-divergence about 1.2. These are due to sequences whose terms all start with the digit $1$, such as sequences that expresses the terms in binary notation (e.g. OEIS sequence A035526) for which $b_d(i) = 1$ if $i=1$ and $0$ otherwise (a distribution we will denote by $\delta_9$) as the corresponding KL-divergence is $\log\left(\frac{1}{\log_{10}(2)}\right) = \log(\log_2(10)) \approx 1.2005$.
Fig. \ref{fig:KS} shows the KS statistic which shows a similar behavior to Fig. \ref{fig:KL}.

We also computed the Wasserstein distance $WD(b_d,b)$ between $b_d$ and $b$ for these sequences. As the Wasserstein distance take into account a permutation of the digits (i.e. the Wasserstein distance does not change if the values of $b_d$ or $b$ are permuted), $WD(b_d,b) \leq WD(\delta_{9},b) = \frac{2(1-\log_{10}(2))}{9} \approx 0.1553$, the plot in Fig. \ref{fig:WD} shows relatively smaller values.

Finally, in Fig. \ref{fig:TV} we show the total variation distance of $b_d$ and $b$ which is similar to Fig. \ref{fig:KS}. 
There are some sequences (i.e. \href{https://oeis.org/A000038}{A000038}) which is all zero except for a single term which has a total variation distance close to $1$.

These figures show that $b_d$ is relatively close to $b$ for many sequences which implies that they adhere to Benford's law.

\subsection{Checking for Taylor's law}
To check for Taylor's law, we compute the following quantities for each sequence $\{a(n)\}$:
$\mu(n) = \frac{1}{n}\sum_{i=1}^n a(i)$ and $v(n) = \frac{1}{n-1}\sum_{i=1}^n (a(i) - \mu(i))^2$ with $v(1) = 0$. 

Note that we use the sample variance as we interpret the sequence as samples from an experimental process. Using the population variance instead ($\frac{1}{n}\sum_{i=1}^n (a(i)-\mu(i))^2 = (n-1)v(n)/n$) gives very similar results as the total number of terms for each sequence is relatively large.

We fitted $\log v$ against $\log \mu$ with a linear regressor to obtain the following features: slope $s$, intercept $b$ and 
correlation coefficient $r$. When $r$ is close to $1$, Taylor's law (Eq. \ref{eqn:taylor}) is closely satisfied with slope $s = T_b$ and the intercept $b = \log(T_a)$.

\begin{figure}[htbp]
\centerline{\includegraphics[width=5in]{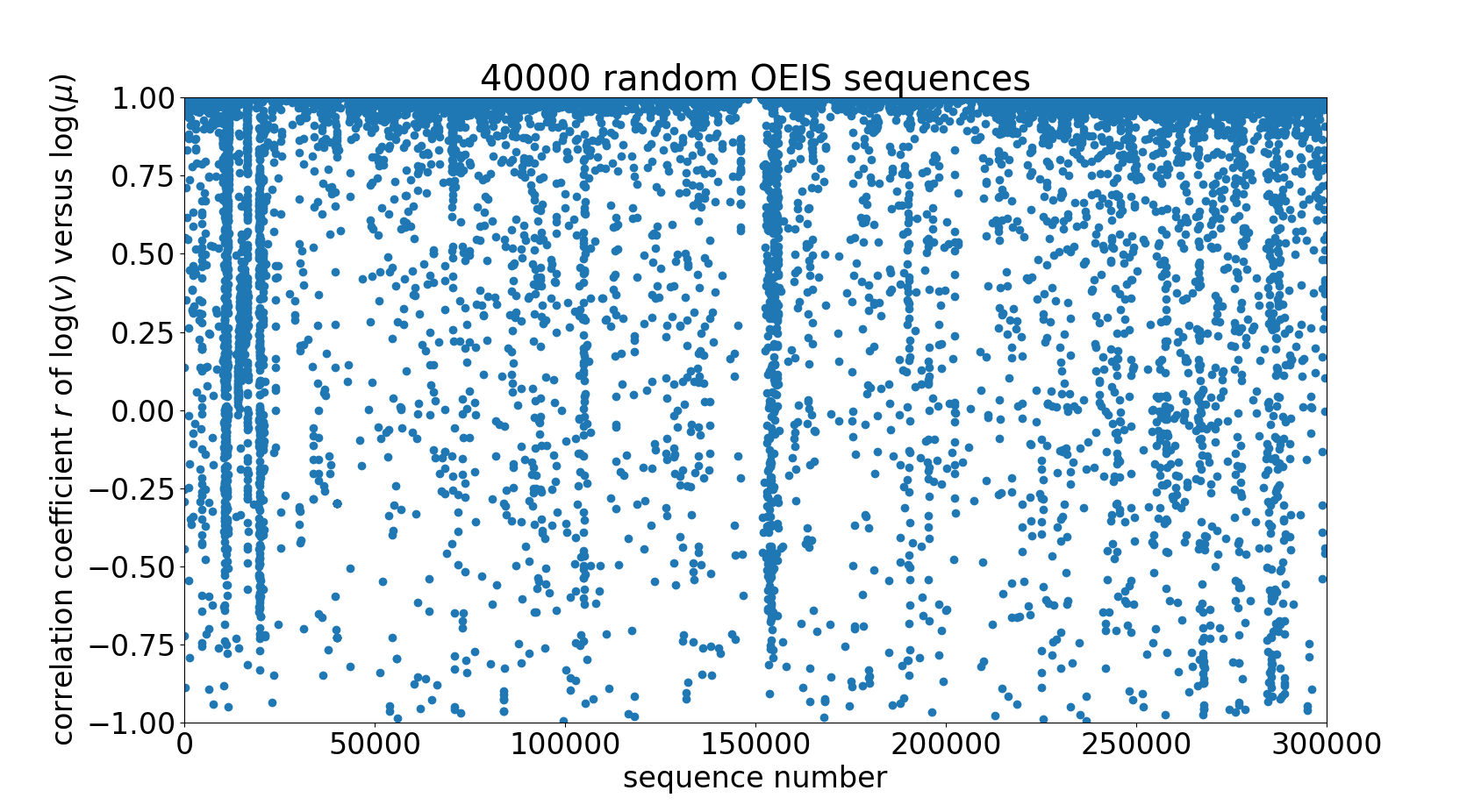}}
\caption{Correlation coefficient $r$ of $\log(v)$ versus $\log(\mu)$.}
\label{fig:cc}
\end{figure}

Figure \ref{fig:cc} shows the correlation coefficient $r$ against the sequences number. We see that for many (but not all) sequences the correlation coefficient $r$ is close to $1$. We also notice sequences where $r$ is negative, indicating a negative correlation. In this case $s$ is negative, corresponding to a negative exponent $T_b$ in Taylor's law. This is quite different from the original form of Taylor's law where $T_b >0$ and observed in general data sets \cite{Xiao2015,Reuman2017,Cohen2016,Demers2018}.

\begin{figure}[htbp]
\centerline{\includegraphics[width=5in]{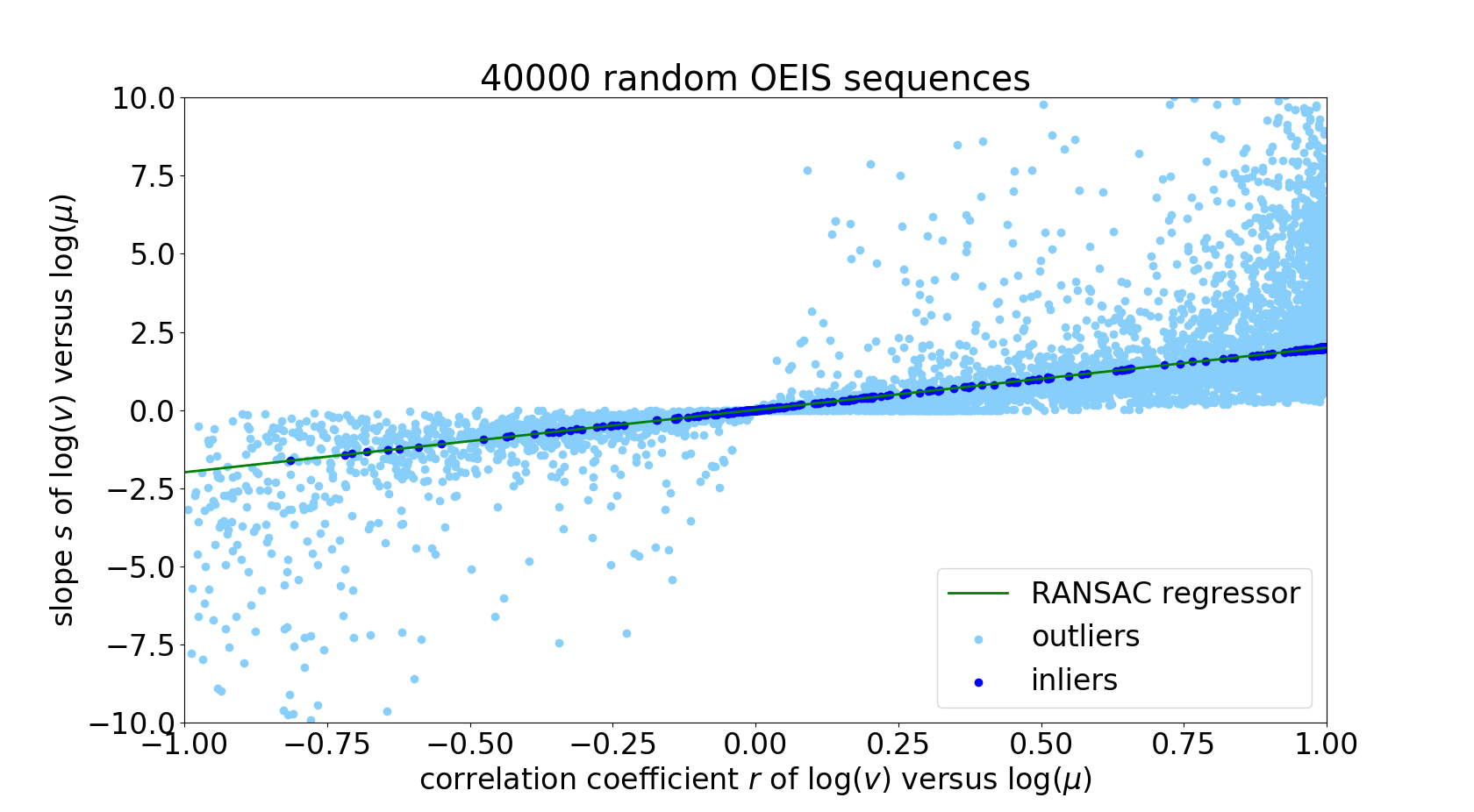}}
\caption{Slope $s$ vs correlation coefficient $r$. The inliers (about half of the sequences) matches the regressor line with slope $2.001$, intercept = $0.003$ with a correlation coefficient equal to $0.999$.}
\label{fig:slope-vs-r}
\end{figure}

Figure \ref{fig:slope-vs-r} shows $r$ plotted against $s$, along with a regressor derived from the RANSAC algorithm \cite{Fischler1987} with a slope of approximately $2$. Since $\frac{s}{r} = \frac{Sl_v}{Sl_\mu}$, where $Sl_v$ and $Sl_\mu$ are the standard deviation of $\log(v)$ and $\log(\mu)$ respectively, this implies that for many sequences $Sl_v \approx 2Sl_\mu$. This is further accentuated by the inliers in Fig. \ref{fig:slope-vs-r}, which represents about $50\%$ of the sequences considered, which matches the regressor line with slope $2$ with a very high correlation coefficient of $0.999$.

For each sequence, we also compute $p_z$ which is the proportion of terms that are positive resulting in a total of $14$ features:
$s$, $b$, $r$ and $p_z$, and $b_d(i)$ for $i=0,\cdots 9$. The feature $b_d(0)$ denotes the proportion of zero terms in the sequence. Note that $p_z$ and the features $b_d(i)$ related to Benford's law is invariant under permutation of the terms of the sequence, whereas the features related to Taylor's law ($s$, $b$ and $r$) do not and we think that both types of features are necessary to properly classify a sequence.

\section{Classifiers for identifying OEIS sequences}
\label{sec:classify}
The above results show that many, but not all sequences satisfy to some degree Benford's law (BL) and Taylor's law (TL), suggesting that BL and TL could be used to identify whether a sequence would be of interest to OEIS editors and users.
For instance, we could argue that  if  $s \approx 2r$, then the sequence is a candidate for inclusion in OEIS. To test this idea we generated approximately 40,000 sequences of 2000 random integers and calculated the $14$ features for these random sequences as well. We add these to the OEIS sequences to obtain a dataset of features from 80,000 sequences and randomly choose 70,000 for training and 10,000 for testing. We implemented a random forest classifier \cite{Breiman2001} with 665 trees and other parameters obtained via hyperparameter optimization.
Preprocessing the data with a Principal Component Analysis, we were able to obtain the following performance metrics in distinguishing OEIS sequences from random sequences: accuracy: 0.999, precision: 0.9984, recall: 0.9996, F$_1$ score: 0.9990.
 
On the other hand, even though it appears relatively easy to distinguish OEIS sequences from random sequences, the complement of OEIS is hard to define precisely. In fact, almost any integer sequence can be submitted to the OEIS and included if the editors deemed it interesting mathematically.  Furthermore, a small perturbation to any sequence in the OEIS will unlikely be in the database, but this will not be detected by this classifier. But given the sequences in the database so far we could draw the conclusion that such interesting sequences tend to satisfy BL and TL (or at least distinguishable by the features derived from BL and TL). The purpose of the next sections is to see if the parameters derived from the sequences to test for adherence to BL and TL can be used to further categorize sequences within OEIS. 

\section{Classifiers for identifying keywords in OEIS sequences}

We first identify the following labels for each sequence.

\subsection{Sequence labels}

We note for each sequence the absence or presence of the following OEIS keywords:
`nice', `core', `easy', `mult'. They describe sequences that are ``nice'', important, easy to compute and multiplicative (in the number theory sense) respectively.
In addition we added the keywords `prime', `binomial', `palindrome'\footnote{A number is called a {\em palindrome} if it is the same when read left to right or right to left. Examples include the numbers 1348431 and 9889. When the base is not specified, the number is assumed to be written in base 10.} if these words appear in the title or in the comments section of the sequence in the OEIS database. We also added a keyword 'other' to denote the absence of any of the above keywords. Thus we have total of $8$ labels for each sequence. Each sequence can have more than one label.

We train different types of classifiers to analyze the dataset. A total of 35000 sequences will be use for training and validation.
The test set consists of 5000 sequences. Preprocessing based on statistics of the training set are applied to normalize the training and test set. Note that these classifiers are not determining whether the sequences satisfy Benford's Law and Taylor's Law (and we have seen that some sequences don't), but whether the features derived from these laws are useful in classification based on the keywords.

\subsection{Neural network}
The neural network has 6 dense layers with 933 neurons and about 140,000 trainable parameters, and using ReLu activation functions, except the output layer which uses a sigmoid activation function. A dropout layer with probability 0.25 is inserted after each input and hidden layer.
We train this neural network for 40000 epochs with a batch size of 32. 
We use 31500 sequences for training in each epoch and 3500 sequences for validation.

\subsection{Random forest ensemble classifier}
We will consider 2 types of ensemble classifiers: The random forest classifier  and the extra trees ensemble classifier.
For both these classifiers, the hyperopt-sklearn module is used to tune the hyperparameters. 
A standard scaling normalizes the data based on the variance and mean of the training set.
The random forest consists of 744 trees, and all features have similar Gini importance. 

\subsection{Extra trees ensemble classifier}
The extra trees (extremely randomized trees) classifier \cite{Geurts2006} is a generalization and an improvement of the random forest classifier. 
The number of trees is 1059 and again all features have similar Gini importance. 

\subsection{Baseline classifier}
As a baseline, we also construct a random classifier, where each predicted label is chosen with a probability derived from the training set.

Since some labels occur much more frequently than other labels, the (subset) accuracy for such a unbalanced problem is generally not the best metric \cite{Valverde-Albacete2014}, and therefore as in Section \ref{sec:classify} we will also compute the precision, recall and F$_1$-score for each classifier.

\section{Experimental results}
\begin{figure}[htbp]
\centerline{\includegraphics[width=5in]{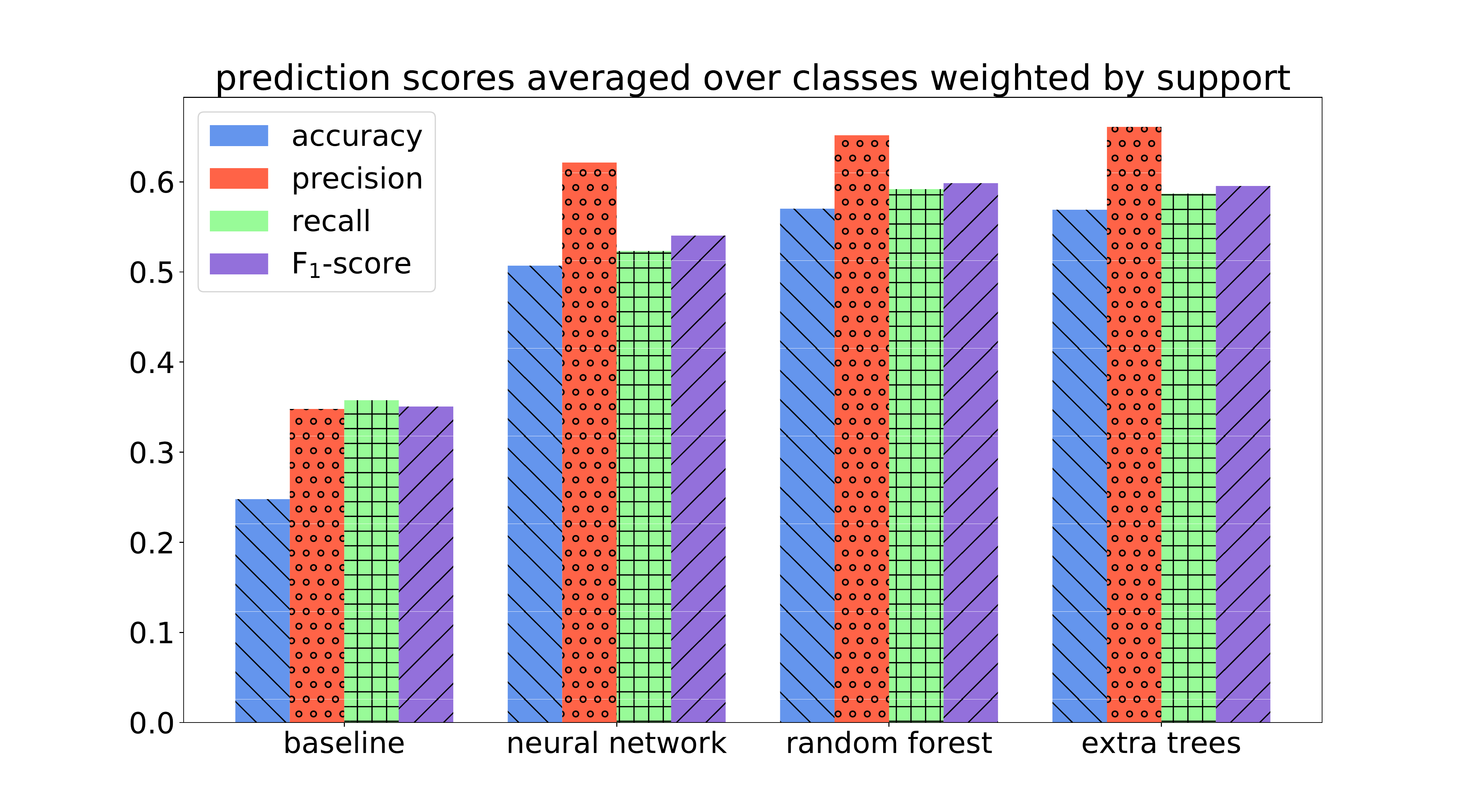}}
\caption{Accuracy, precision, recall and F$_1$ score of 3 classifiers compared with the random classifier.}
\label{fig:performance}
\end{figure}

\begin{figure}[htbp]
\begin{subfigure}{0.5\textwidth}
\centerline{\includegraphics[width=3in]{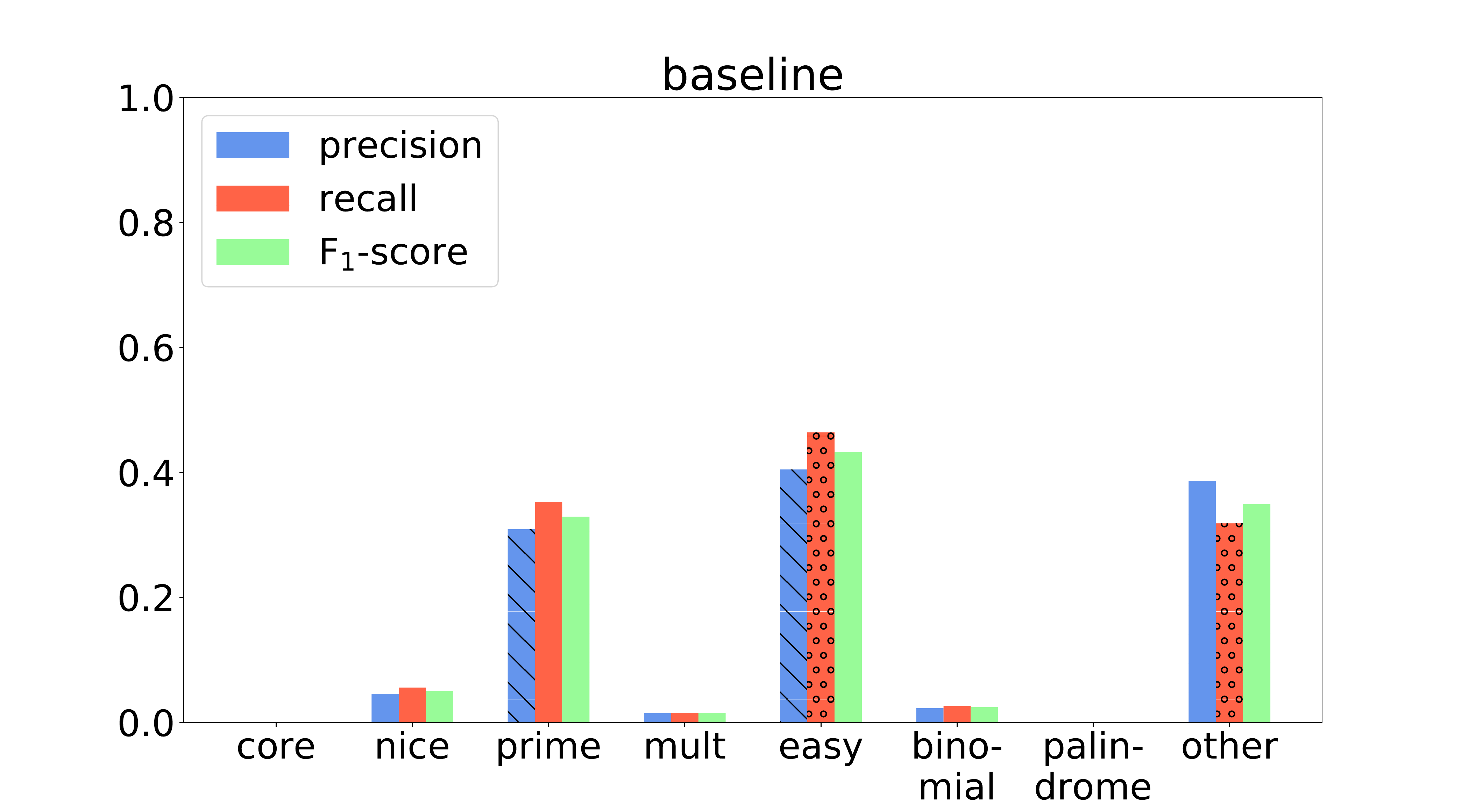}}
\caption{Baseline classifier.}
\label{fig:rg}
\end{subfigure}
\begin{subfigure}{0.5\textwidth}
\centerline{\includegraphics[width=3in]{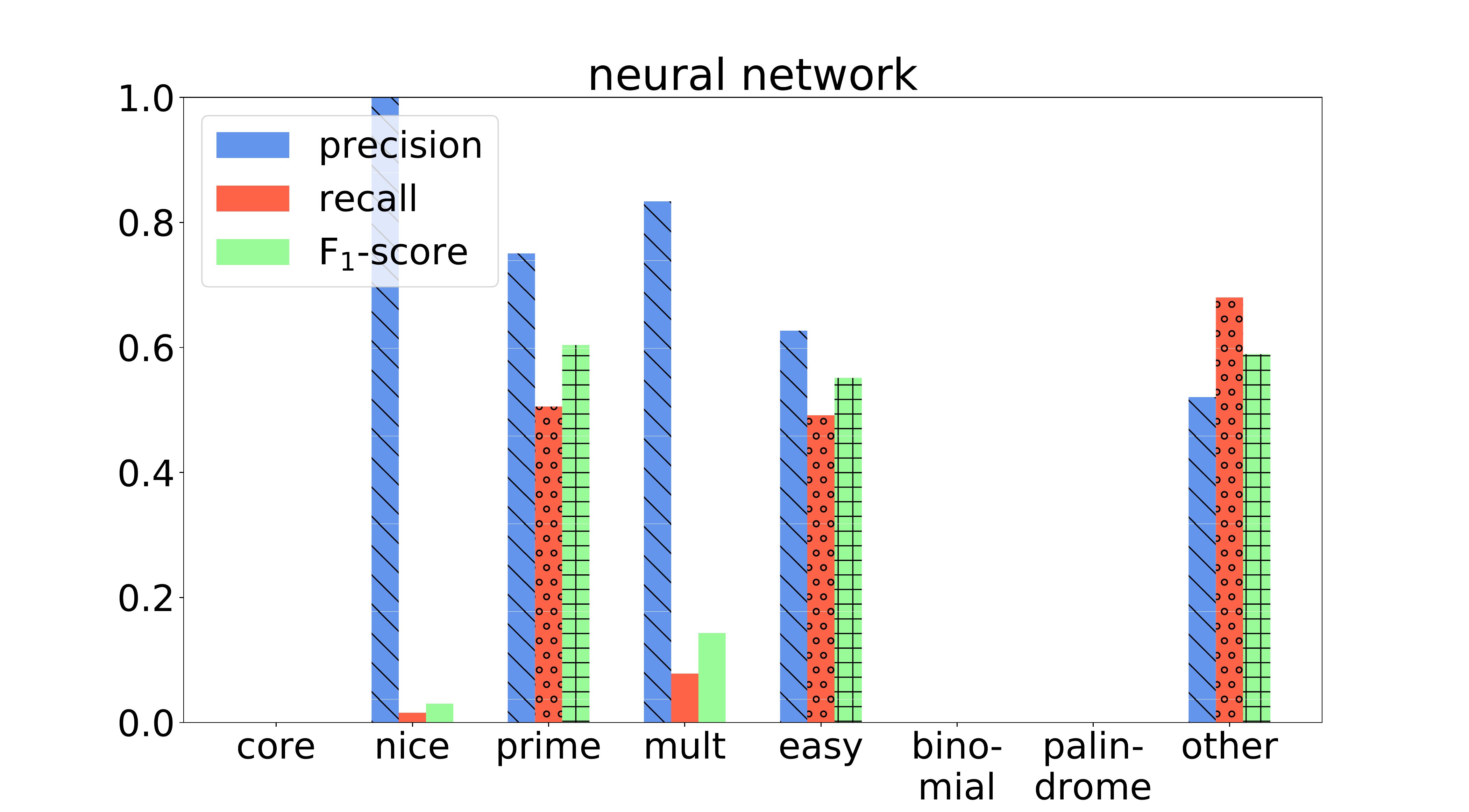}}
\caption{Neural network classifier.}
\label{fig:nn}
\end{subfigure}

\bigskip
\begin{subfigure}{0.5\textwidth}
\centerline{\includegraphics[width=3in]{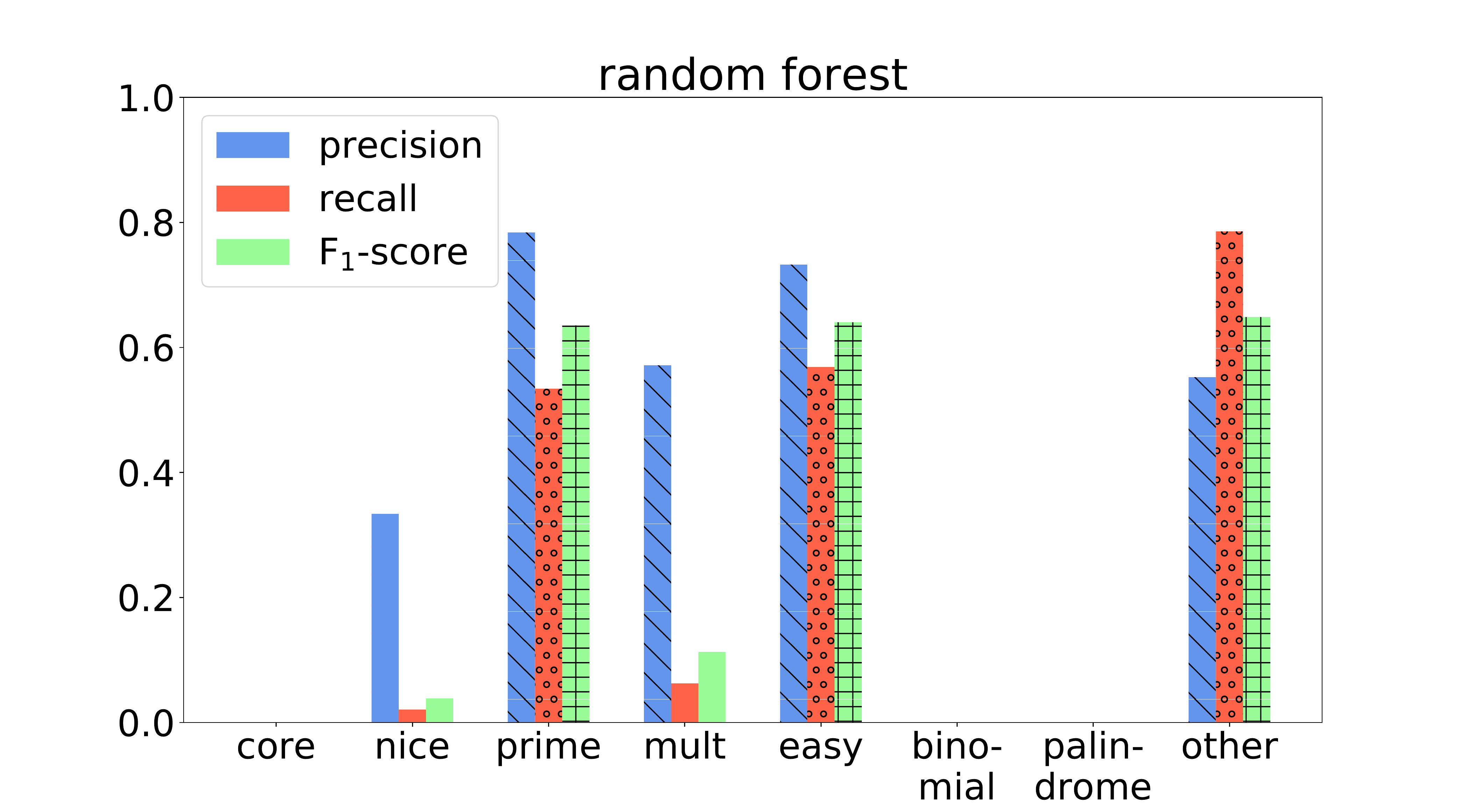}}
\caption{Random forest classifier.}
\label{fig:rf}
\end{subfigure}
\begin{subfigure}{0.5\textwidth}
\centerline{\includegraphics[width=3in]{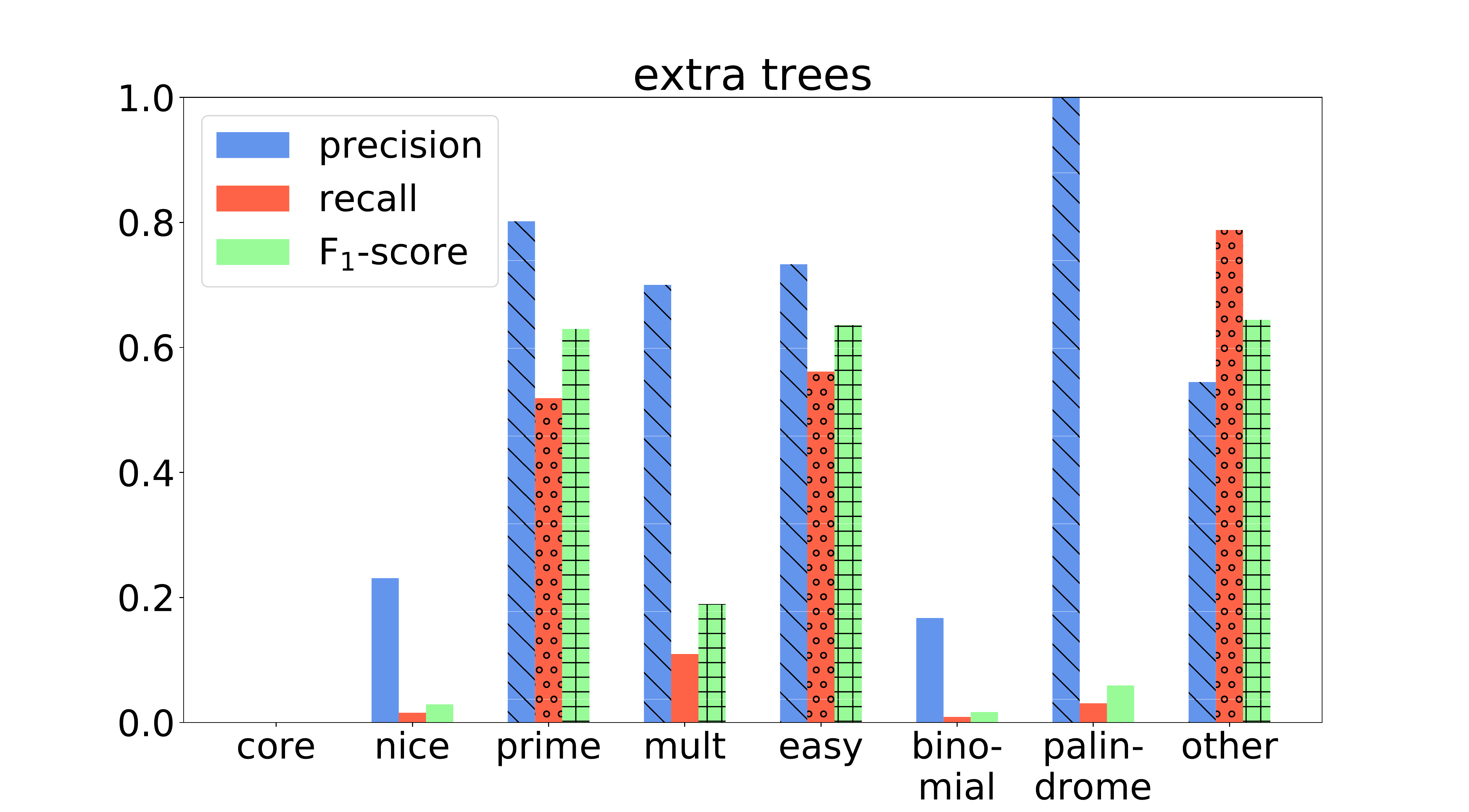}}
\caption{Extra trees classifier.}
\label{fig:et}
\end{subfigure}
\caption{Performance (precision, recall and F$_{1}$ score) of various classifiers on each label.}
\label{fig:performanceclass}
\end{figure}

The performance of these various models in predicting each label is shown in Figure \ref{fig:performance}, where we plotted the (subset) accuracy, precision, recall and F$_1$-score of each model\footnote{Precision, recall and F$_1$ score is set to $0$ when it is undefined (i.e. the denominator is $0$).}. Since each sequence can have multiple labels, these quantities are computed for each label and are averaged among the labels weighted by their support. In Figure \ref{fig:performanceclass} we plot these quantities for each of the labels for each of the models. We find that the extra trees ensemble classifier performs the best, followed by a random forest classifier,  and then a deep neural network. All of them performed better than the baseline classifier. Furthermore, they all had problems classifying labels that are not well supported: `nice', `core', `palindrome', `binomial' and `mult', a well known problem of multilabel data sets that are not balanced.
Note that for the label "palindrome" the extra trees classifier has a nontrivial recall and precision unlike the other classifiers for which these quantities are either $0$ or undefined.
The neural network classifier has perfect precision (no false positives and at least one true positive) for `nice' sequences and the extra trees classifier has perfect precision of `palindrome' sequences.
Note also that the scores for the `mult' labels are significantly higher for the 3 classifiers versus the baseline classifier, suggesting that multiplicative sequences can be detected using these empirical laws. Can this conclusion be a consequence of the definition of multiplicative?

The performance of the models are not stellar, but it is better than the baseline classifier. Part of this could be due to the fact that a small number of features (14) are used in the classifiers. As the OEIS database consists of sequences that people have submitted and the editors approved, it is biased towards sequences which people have found to be interesting or useful mathematically. This means that the ability to predict sequences with the "prime" label does not mean it was able to classify all prime-related sequences, but merely that it was able to classify prime sequences that are interesting or relevant. Furthermore, labels such as `nice' and `easy' are subjective and can vary depending on the person who assigned the label, and such issues are common in classification tasks such as sentiment analysis \cite{Jain2016}. 

In some cases the models were able to find related sequences. For instance sequence A059260 was in the training set which included the labels `nice' and `binomial'. The sequence A059259 which was in the test set is a related sequence which enumerated the triangle of terms in a different order and the extra trees classifier also predicted the labels `nice' and `binomial' even though these labels were not assigned to A059259.
The extra trees classifier predicted the label `binomial' for sequence A080575 which is appropriate since it list the triangle of multinomial coefficients.
Similarly, sequence A182009 which is an approximation (and is almost identical to) of sequence A033810 (which is in the training set), the extra trees classifier also predicted the labels of A033810 for sequence A182009.

\section{Conclusions}
It is difficult to define what constitute interesting mathematical structures. If it is parsimony of representation, then perhaps something like Kolmogorov complexity would be an appropriate metric. We take the alternative approach that since many data sets follow empirical laws, these laws are harbingers of interesting integer sequences. In our experiments, we use OEIS as a proxy of what mathematical sequences are of interest to us.
The experimental results point to the possibility of classifying interesting or relevant integer sequences using derived parameters based on empirical laws. They seem to indicate that we can differentiate mathematically interesting OEIS sequences from random sequences and that certain characteristics of these sequences can be identified solely based on the numbers in the sequence.  A possible explanation for this is that the empirical laws are capturing inherent salient properties of numerical data that are interesting or important to study. Future work include training a deeper network, and adding other features perhaps from other empirical laws (such as Zipf's law \cite{Zipf1935}\footnote{Since most sequences have distinct terms, in order to test Zipf's law some methods of grouping the terms into disjoint sets are needed.} which is a general form of Benford's law) to see if the performance improves. Some other ideas to consider include looking at the mean and variance of $n$-th order differences and how well a sequence fits a polynomial equation or a linear recurrence relationship. An interesting open question to investigate is why so many OEIS sequences follows $s = 2r$ so closely.

\section*{Acknowledgements}
We would like to thank Neil Sloane for introducing us to Taylor's law and Benford's law and for his encouragement and helpful comments to a draft of this manuscript.


\begin{thebibliography}{18}
\providecommand{\natexlab}[1]{#1}
\providecommand{\url}[1]{\texttt{#1}}
\expandafter\ifx\csname urlstyle\endcsname\relax
  \providecommand{\doi}[1]{doi: #1}\else
  \providecommand{\doi}{doi: \begingroup \urlstyle{rm}\Url}\fi

\bibitem[Akritidis and Bozanis(2013)]{Akritidis2013}
Leonidas Akritidis and Panayiotis Bozanis.
\newblock A {Supervised} {Machine} {Learning} {Classification} {Algorithm} for
  {Research} {Articles}.
\newblock In \emph{Proceedings of the 28th {Annual} {ACM} {Symposium} on
  {Applied} {Computing}}, {SAC} '13, pages 115--120, New York, NY, USA, 2013.
  ACM.

\bibitem[Kushman et~al.(2014)Kushman, Artzi, Zettlemoyer, and
  Barzilay]{Kushman2014}
Nate Kushman, Yoav Artzi, Luke Zettlemoyer, and Regina Barzilay.
\newblock Learning to automatically solve algebra word problems.
\newblock In \emph{Proceedings of the 52nd Annual Meeting of the Association
  for Computational Linguistics (Volume 1: Long Papers)}, volume~1, pages
  271--281, 2014.

\bibitem[{The OEIS Foundation Inc.}(1996-present)]{oeis}
{The OEIS Foundation Inc.}
\newblock The on-line encyclopedia of integer sequences, 1996-present.
\newblock URL \url{https://oeis.org/}.
\newblock Founded in 1964 by N. J. A. Sloane.

\bibitem[Moore(1965)]{Moore1965b}
G~Moore.
\newblock Cramming more components onto integrated circuits.
\newblock \emph{Electronics}, 38\penalty0 (8):\penalty0 114--117, 1965.

\bibitem[Benford(1938)]{Benford1938}
Frank Benford.
\newblock The law of anomalous numbers.
\newblock \emph{Proceedings of the American philosophical society}, 78\penalty0 (4):\penalty0 551--572, 1938.

\bibitem[Newcomb(1881)]{Newcomb1881}
Simon Newcomb.
\newblock Note on the frequency of use of the different digits in natural
  numbers.
\newblock \emph{American Journal of Mathematics}, 4\penalty0 (1):\penalty0
  39--40, 1881.

\bibitem[Hürlimann(2009)]{Huerlimann2009}
Werner Hürlimann.
\newblock Generalizing {Benford}'s {Law} {Using} {Power} {Laws}: {Application}
  to {Integer} {Sequences}.
\newblock \emph{International Journal of Mathematics and Mathematical
  Sciences}, 2009:\penalty0 1--10, 2009.

\bibitem[Taylor(1961)]{Taylor1961}
LR~Taylor.
\newblock Aggregation, variance and the mean.
\newblock \emph{Nature}, 189\penalty0 (4766):\penalty0 732--735, 1961.

\bibitem[Xiao et~al.(2015)Xiao, Locey, and White]{Xiao2015}
Xiao Xiao, Kenneth~J Locey, and Ethan~P White.
\newblock A process-independent explanation for the general form of {T}aylor’s
  law.
\newblock \emph{The American Naturalist}, 186\penalty0 (2):\penalty0 E51--E60,
  2015.

\bibitem[Reuman et~al.(2017)Reuman, Zhao, Sheppard, Reid, and
  Cohen]{Reuman2017}
Daniel~C. Reuman, Lei Zhao, Lawrence~W. Sheppard, Philip~C. Reid, and Joel~E.
  Cohen.
\newblock Synchrony affects {Taylor}’s law in theory and data.
\newblock \emph{Proceedings of the National Academy of Sciences}, page
  201703593, May 2017.

\bibitem[Cohen(2016)]{Cohen2016}
Joel~E. Cohen.
\newblock Statistics of {Primes} (and {Probably} {Twin} {Primes}) {Satisfy}
  {Taylor}'s {Law} from {Ecology}.
\newblock \emph{The American Statistician}, 70\penalty0 (4):\penalty0 399--404,
  October 2016.

\bibitem[Demers(2018)]{Demers2018}
Simon Demers.
\newblock Taylor's {Law} {Holds} for {Finite} {OEIS} {Integer} {Sequences} and
  {Binomial} {Coefficients}.
\newblock \emph{The American Statistician}, January 2018.

\bibitem[Fischler and Bolles(1987)]{Fischler1987}
Martin~A Fischler and Robert~C Bolles.
\newblock Random sample consensus: a paradigm for model fitting with
  applications to image analysis and automated cartography.
\newblock In \emph{Readings in computer vision}, pages 726--740. Elsevier,
  1987.

\bibitem[Breiman(2001)]{Breiman2001}
Leo Breiman.
\newblock Random forests.
\newblock \emph{Machine learning}, 45\penalty0 (1):\penalty0 5--32, 2001.

\bibitem[Geurts et~al.(2006)Geurts, Ernst, and Wehenkel]{Geurts2006}
Pierre Geurts, Damien Ernst, and Louis Wehenkel.
\newblock Extremely randomized trees.
\newblock \emph{Machine learning}, 63\penalty0 (1):\penalty0 3--42, 2006.

\bibitem[Valverde-Albacete and Pel{\'a}ez-Moreno(2014)]{Valverde-Albacete2014}
Francisco~J Valverde-Albacete and Carmen Pel{\'a}ez-Moreno.
\newblock 100\% classification accuracy considered harmful: The normalized
  information transfer factor explains the accuracy paradox.
\newblock \emph{PloS one}, 9\penalty0 (1):\penalty0 e84217, 2014.

\bibitem[Jain and Dandannavar(2016)]{Jain2016}
Anuja~P Jain and Padma Dandannavar.
\newblock Application of machine learning techniques to sentiment analysis.
\newblock In \emph{Applied and Theoretical Computing and Communication
  Technology (iCATccT), 2016 2nd International Conference on}, pages 628--632.
  IEEE, 2016.

\bibitem[Zipf(1935)]{Zipf1935}
George~Kingsley Zipf.
\newblock The psycho-biology of language.
\newblock 1935.

\end{thebibliography}
\end{document}